\pdfoutput=1
\documentclass{bmvc2k}
\usepackage{multirow}
\usepackage{dingbat}
\usepackage{caption}
\usepackage[export]{adjustbox}
\usepackage{verbatim}
\usepackage{wrapfig}
\usepackage{bbm}
\usepackage{hyperref}
\usepackage[font=small,labelfont=bf]{caption}

\title{Improved Bilinear Pooling with CNNs}

\addauthor{Tsung-Yu Lin}{tsyungyulin@cs.umass.edu}{1}
\addauthor{Subhransu Maji}{smaji@cs.umass.edu}{1}

\addinstitution{
 University of Massachusetts, Amherst\\
 Amherst, MA 01003
}

\runninghead{Lin and Maji}{Improved Bilinear Pooling with CNNs}

\def\eg{\emph{e.g}\bmvaOneDot}
\def\ie{\emph{i.e}\bmvaOneDot}

\def\etal{\emph{et al}\bmvaOneDot}
\def\sqA{$A^{1/2}$}

\begin{document}
\maketitle

\begin{abstract}
Bilinear pooling of Convolutional Neural Network (CNN) features~\cite{lin2015bilinear,lin2017bilinear}, and their compact variants~\cite{gao16compact}, have been shown to be effective at fine-grained recognition, scene categorization, texture recognition, and visual question-answering tasks among others. 
The resulting representation captures second-order statistics of convolutional features in a translationally invariant manner. 
In this paper we investigate various ways of normalizing these statistics to improve their representation power. In particular we find that the \emph{matrix square-root} normalization offers significant improvements and outperforms alternative schemes such as the \emph{matrix logarithm} normalization when combined with \emph{elementwise square-root} and $\ell_2$ normalization. 
This improves the accuracy by \textbf{2-3\%} on a range of fine-grained recognition datasets leading to a new state of the art.

We also investigate how the accuracy of matrix function computations effect network training and evaluation. In particular we compare against a technique for estimating matrix square-root gradients via solving a Lyapunov equation that is more numerically accurate than computing gradients via a Singular Value Decomposition~(SVD). We find that while SVD gradients are numerically inaccurate the overall effect on the final accuracy is negligible once boundary cases are handled carefully. We present an alternative scheme for computing gradients that is faster and yet it offers improvements over the baseline model. Finally we show that the matrix square-root computed approximately using a few Newton iterations is just as accurate for the classification task but allows an order-of-magnitude faster GPU implementation compared to SVD decomposition.

\end{abstract}

\section{Introduction}
\label{sec:intro}
Convolutional Neural Networks (CNNs) trained on large-scale image classification tasks have emerged as powerful general-purpose feature extractors in recent years. Representations using the activations of layers of such networks have been shown to be effective for tasks ranging from object detection, semantic segmentation, texture recognition, to fine-grained recognition of object categories. While most techniques extract activations from the penultimate layers of the network, a different line of work has considered aggregating higher-order statistics of CNN activations. Examples include the VLAD~\citep{jegou10aggregating,Arandjelovic16}, Fisher vector~\cite{perronnin07fisher,cimpoi2015deep} and bilinear (second-order) pooling~\cite{carreira2012semantic,lin2015bilinear,ionescu2015matrix}. These techniques are inspired by classical texture representations which were build on hand-designed filter banks (\eg, wavelets or steerable pyramids) or SIFT features~\cite{lowe99object}.
Second-order aggregation of CNN activations not only provides significant improvements over classical representations but also is more effective than those built using first-order aggregation (\eg, sum or max) for tasks such as texture synthesis~\cite{gatys2015texture}, style transfer~\cite{Gatys_2016_CVPR}, image classification tasks~\cite{cimpoi2015deep,lin2015bilinear,lin16visualizing}. 
In particular the Bilinear CNN (B-CNN)~\cite{lin2015bilinear,lin2017bilinear} has emerged as a state-of-the-art network architecture for texture and fine-grained recognition.

In this paper we investigate the effect of \emph{feature normalization} in detail and study its impact on the accuracy of the B-CNN. In particular we investigate a class of matrix functions that scale the spectrum (eigenvalues) of the covariance matrix obtained after bilinear pooling. One such normalization is the \emph{matrix-logarithm} function defined for Symmetric Positive Definite (SPD) matrices. It maps the Riemannian manifold of SPD matrices to a Euclidean space that preserves the geodesic distance between elements in the underlying manifold. Empirically, the O2P approach~\cite{carreira2012semantic} showed that this mapping results in better performance when used with linear classifiers. A follow-up work by Ionescu \etal~\cite{ionescu2015matrix} extended the approach using deep features and showed similar improvements for semantic segmentation. Huang and Gool~\cite{huang2017riemannian} recently applied this technique for image classification using second-order features.
 However, computing the logarithm of a matrix and its gradient is expensive and numerically unstable on modern GPUs. Hence, normalization using matrix functions remains largely unexplored for training state-of-the-art networks.

The matrix-logarithm mapping applies a non-linear scaling to the eigenvalues of the matrix. 
This is because the logarithm of a SPD matrix $A$ with a Singular Value Decomposition~(SVD) of $A = U\Sigma U^{T}$ is given by $\log(A)=U\log(\Sigma)U^{T}$. Here $\log(\Sigma)$ applies logarithm to the diagonal elements of $\Sigma$ in an elementwise manner. 
While elementwise square-root compensates for "burstiness" of visual words in a bag-of-visual-words model~\cite{perronnin10improving}, spectral normalization is appropriate for covariance matrices since correlated features contribute to the eigenvalues in a similar manner. 
Based on this observation we consider schemes that apply a power normalization to the spectrum of the matrix. For example, the matrix square-root $A^{1/2}=U\Sigma^{1/2}U^{T}$ applies elementwise square-root to the diagonal elements of $\Sigma$. 
Other matrix powers can be defined similarly. Our experiments show that matrix-normalization offers complementary benefits to elementwise normalization schemes. Empirically, the combination of matrix square-root and elementwise square-root works the best and leads to a \textbf{2-3\%} improvement in accuracy over the baseline B-CNN approach on three fine-grained recognition datasets. Moreover, the matrix square-root can be more efficiently computed than matrix logarithm using variants of Newton iterations on the GPU.

The gradients of the matrix functions can also be derived using the SVD.  However, since the SVD is ill-conditioned when the matrix has eigenvalues that are numerically close. Prior work of Ionescu \etal~\cite{ionescu2015matrix} has noted this problem and suggested that double precision computations were necessary for training. Fortunately, for the case of matrix square-root one can avoid the SVD decomposition entirely and compute the gradients by solving a Lyapunov equation. This allows us to systematically evaluate the effect of numerical precision of gradients on the accuracy of the final model. In particular we find that training using Lyapunov gradients indeed leads to more accurate models. A simpler but less accurate scheme that ignores the matrix normalization entirely during fine-tuning also improves over the baseline model. Finally, we propose techniques to improve the efficiency of matrix square-root computation on GPUs. Computing the square-root via a SVD on GPUs is a bottleneck in the network evaluation. Instead we explore variants of Newton iterations for computing the square-root. 
In particular we show that few iterations of a modified Denman-Beavers iterations are sufficient for classification needs. Moreover, these iterations only involve matrix multiplications, are easily parallelizable, and are an order-of-magnitude faster than SVD-based approach for square-root computation.


\section{Method and related work}\label{sec:method}
\begin{figure}
\centering
\includegraphics[width=\linewidth]{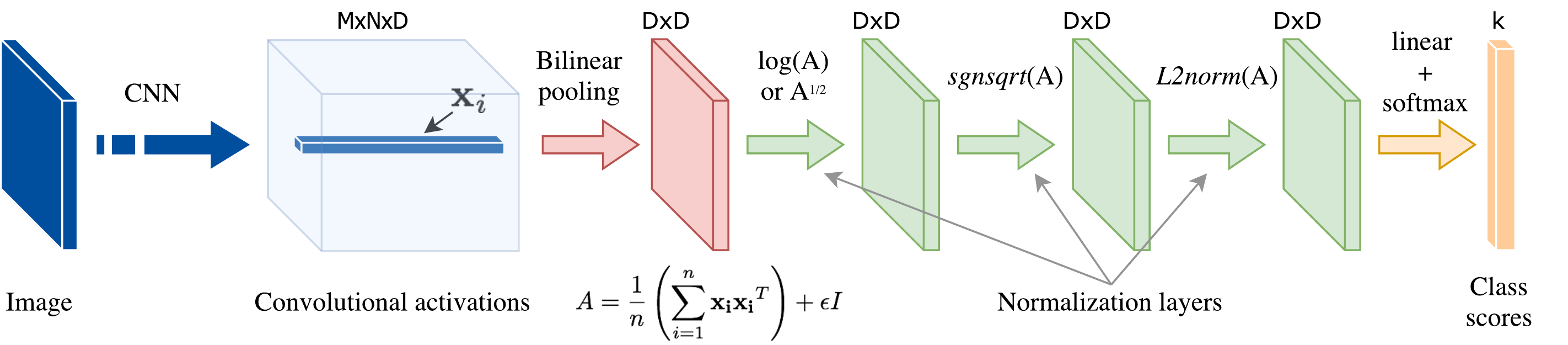}
\vspace{0.01in}
\caption{\label{fig:arch} \textbf{Improved B-CNN architecture} with a $\log(A)$ or $A^{1/2}$, signed square-root, and $\ell_2$ normalization layers added after the bilinear pooling of CNN activations.}
\vspace{-0.1in}
\end{figure}

Our method builds on the B-CNN architecture proposed in our earlier work~\cite{lin2015bilinear,lin2017bilinear}. We review the architecture and describe the improvements based on matrix-function layers in Section~\ref{sec:bcnn}. We then describe different ways these layers can be implemented that tradeoff numerical precision and accuracy in Section~\ref{sec:matrix-functions}.

\subsection{The B-CNN architecture}\label{sec:bcnn}
The B-CNN architecture employs bilinear pooling to aggregate the locationwise outer-product of two features by global averaging. This results in a covariance matrix which captures the pairwise interactions between the two features. In this work we analyze the case when the two features are identical which results in a symmetric positive semi-definite matrix. Following the terminology of our earlier work~\cite{lin2015bilinear,lin2017bilinear} these are symmetric B-CNNs and are identical to the Second-Order Pooling (O2P)~\cite{carreira2012semantic} popularized for semantic segmentation. 

The network architecture is illustrated in Figure~\ref{fig:arch}. Given an image a CNN is used to extract a set of features $\mathbf{x}_i$ across locations $i=1,2,\ldots,n$. The bilinear (second-order) pooling extracts the second-order statistics and adds a small positive value $\epsilon$ to the diagonal resulting in matrix $A$ given by:
\begin{equation}
A = \frac{1}{n}\left(\sum_{i=1}^{n} \mathbf{x_i}\mathbf{x_i}^T\right) + \epsilon I
\end{equation}
Given a feature $\mathbf{x_i}$ of $d$ dimensions the matrix $A$ is of size $d \times d$.
Our earlier work~\cite{lin2015bilinear} showed that normalization of the matrix $A$ is critical for good performance. In particular, elementwise signed square-root $(x \leftarrow sign(x)\sqrt{|x|})$ and $\ell_2$ normalization is applied to the matrix $A$ before it is plugged into linear classifiers. Both the pooling and normalization steps are efficient and piecewise differentiable and hence the entire network can be trained in an end-to-end manner by back-propagating the gradients of the objective function (\eg, cross-entropy loss for classification tasks).

\paragraph{Normalization using matrix functions.} The improved B-CNN architecture additionally applies a matrix function normalization to the matrix $A$ after pooling (Figure~\ref{fig:arch}). In particular we consider the matrix logarithm $\log(A)$ originally proposed in the O2P scheme and the matrix power function $A^{p}$ for fractional positive values of $0< p < 1$. Of particular interest is when $p=1/2$ which corresponds to the matrix square-root defined as a matrix $Z$ such that $ZZ=A$. Unlike elementwise transformations, matrix-functions require computations that depend on the entire matrix. Approaches include variants of Newton iterations or via a Singular Value Decomposition (SVD) (described next). 

\subsection{Matrix functions and their gradients}\label{sec:matrix-functions}

\paragraph{Computing matrix functions and their gradients via SVD.} 
Ionescu \etal~\cite{ionescu2015matrix} explore matrix back-propagation for training CNNs and using techniques for computing the derivative of matrix functions (\eg,~\cite{magnus1988matrix}). Given matrix $A$ with a SVD given by $A=U\Sigma U^T$, where the matrix $\Sigma=diag(\sigma_1, \sigma_2, \ldots, \sigma_n)$, the matrix function $f$ can be written as $Z=f(A)=Ug(\Sigma)U^T$, where $g$ is applied to the elements in the diagonal of $\Sigma$. Given the gradient of a scalar loss $L$ with respect to $Z$, the gradient of $L$ with respect to $A$ can be computed as:
\begin{equation}\label{eq:svd-gradients}
\frac{\partial L}{\partial A} = U\Bigg\{ \bigg(K^T \odot \bigg( U^T\frac{\partial L}{\partial U}\bigg) \bigg) + \bigg( \frac{\partial L}{\partial \Sigma}\bigg)_{diag}\Bigg\}U^T.
\end{equation}
Here $\odot$ denotes element-wise matrix multiplication. The matrix $K$ is a skew-symmetric matrix given by $K_{i,j} = 1/(\sigma_i -\sigma_j)\mathbf{I}(i\neq j)$, where $\mathbf{I}(\cdot)$ is the indicator function. The gradients of $L$ with respect to $U$ and $\Sigma$ are:


\begin{equation}
\frac{\partial L}{\partial U} = \bigg\{ \frac{\partial L}{\partial Z} + \bigg(\frac{\partial L}{\partial Z}\bigg)^T\bigg\}Ug(\Sigma),~~~~
\frac{\partial L}{\partial \Sigma} = g'(\Sigma)U^T\frac{\partial L}{\partial Z}U.
\end{equation}
Here $g'(\Sigma) $ is the gradient of the $g$ with respect to $\Sigma$. Since $g$ is applied in an elementwise manner the gradients can be computed easily. For example for the matrix square-root
\begin{equation}
 g'(\Sigma) = diag\left(\frac{1}{2\sqrt{\sigma_1}},  \frac{1}{2\sqrt{\sigma_2}}, \ldots,  \frac{1}{2\sqrt{\sigma_n}}\right).
 \end{equation}
 
\paragraph{Gradients of matrix-square root by solving a Lyapunov equation.} Given a symmetric PSD matrix $A$, with $Z=A^{1/2}$, and a small change $dA$ to the matrix $A$, the change $dZ$ to the matrix $Z$ satisfies the equation:
\begin{equation}
 A^{1/2}dZ + dZA^{1/2}  = dA.
 \end{equation}
  This can be derived by applying the product rule of derivatives to the equation $ZZ = A$. From this one can derive the corresponding chain rule for the loss $L$ as:
\begin{equation}\label{eq:lyap}
  A^{1/2} \left( \frac{\partial L}{\partial A}\right) + \left( \frac{\partial L}{\partial A}\right) A^{1/2}  = \frac{\partial L}{\partial Z}.
\end{equation}
The above is a Lyapunov equation~\cite{lyapunov1992general} which has a closed-form solution given by: 
\begin{equation}\label{eq:lyap_sol}
vec\left( \frac{\partial L}{\partial A}\right) = \left(A^{1/2}\otimes I + I \otimes A^{1/2}\right)^{-1}vec\left(\frac{\partial L}{\partial Z}\right).
\end{equation}
Here $\otimes$ is the Kronecker product and the $vec(\cdot)$ operator unrolls a matrix to a vector. However the above expression can be solved more efficiently using the Bartels-Stewart algorithm~\cite{bartels1972solution}.

\paragraph{Numerical stability and truncated SVD gradients.} While the SVD can be used to compute arbitrary matrix functions in the forward step, computing the gradients using the Equation~\ref{eq:svd-gradients} is problematic when the matrix $A$ has eigenvalues that are close to each other. This stems from the fact that SVD is ill-conditioned in this situation. Adding a small $\epsilon$ to the diagonal of $A$ does not solve this problem either. In practice a truncated SVD gradient where the matrices $K$ and $\Sigma$ are set to zero for indices corresponding to eigenvalues that falls below a threshold $\tau$ works wells. However, even with the truncated SVD we found that the gradient computation results in numerical exceptions. Simply ignoring these cases worked well for fine-tuning when the learning rates were small but their impact on training networks from scratch remains unclear. Lyapunov gradients on the other hand are numerically stable because the inverse in Equation~\ref{eq:lyap_sol} depends on $1/\sigma_{\min}$ where $\sigma_{\min}$ is the smallest eigenvalue of the matrix $A^{1/2}$. Moreover, they can be computed as efficiently as the SVD gradients. In Section~\ref{sec:precision} we evaluate how numerical precision of gradients effects the accuracy.

\paragraph{Newton's method for computing the matrix square-root.} A drawback of SVD based computations of matrix functions is that the computation on GPU is currently poorly supported and sometimes slower than CPU computations. In practice for the networks we consider the time taken for the SVD is comparable to the rest of the network evaluation. For smaller networks this step can become the bottleneck. Instead of computing the matrix square-root accurately, one can instead run a few iterations of a Netwon's method for root finding to the equation $F(Z) = Z^2 - A = 0$. Higam~\cite{higham1997stable} describes a number of variants and analyzes their stability and convergence properties. One such is the Denman-Beavers iterations~\cite{denman1976matrix}. Given $Y_0 = A$ and $Z_0 = I$, where $I$ is the identity matrix, the iteration is defined by
\begin{equation}
Y_{k+1} = \frac{1}{2} (Y_k + Z_k^{-1}),~~
Z_{k+1} = \frac{1}{2} (Z_k + Y_k^{-1}). 
\end{equation}
The matrices $Y_k$ and $Z_k$ converge quadratically to $A^{1/2}$ and $A^{-1/2}$ respectively. In practice about 20 iterations are sufficient. However, these iterations are not GPU friendly either since they require computing matrix inverses which also lack efficient GPU implementations. A slight modification of this equation obtained by replacing the inverses using a single iteration of Newton's method for computing inverses~\cite{ben1965iterative} results in a different update rule given by:
\begin{equation}
Y_{k+1} = \frac{1}{2} Y_k(3I- Z_kY_k),~~
Z_{k+1} = \frac{1}{2} (3I - Z_kY_k)Z_k. 
\end{equation}
However, unlike the Denman-Beavers iterations the above iterations are only locally convergent, \ie, they converge if $||A-I||_2 < 1$. In practice one can scale the matrix $A = \alpha A$ to satisfy this condition~\cite{ben1966note}. These iterations only involve matrix multiplications and are an order-of-magnitude faster than SVD computations on the GPU. Moreover, one can run these iterations for a small number of iterations (even one!) to trade-off accuracy and speed. In Section~\ref{sec:precision} we evaluate the effect of the accuracy of forward computation on the performance of the network. The matrix logarithm can be obtained by iterative scaling and square-root using the identity $\log(A) = 2^k \log(A^{1/2^k})$ and using a Taylor series expansion of the matrix logarithm when the $A^{1/2^k}$ becomes sufficiently close to identity matrix. In this case computing matrix logarithm is inherently slower than computing matrix square-root.

\section{Experiments}
\label{sec:experiments}
We describe the datasets and our experimental setup in Section~\ref{sec:datasets} and Section~\ref{sec:models}.
The accuracy of B-CNNs on three fine-grained classification datasets with various normalization schemes are presented in Section~\ref{sec:norm} with non fine-tuned networks. These results show that the matrix square-root offers complementary benefits to the elementwise normalization proposed in earlier work. We then show results using end-to-end fine-tuning with the normalization layers in Section~\ref{sec:fine-tuning}. We present variants of Newton iterations for computing the matrix square-root and evaluate the impact on the accuracy in Section~\ref{sec:precision}.
We then compare these results to prior work in Section~\ref{sec:prior-work}, especially with other techniques that improve the performance of B-CNNs using ensembles~\cite{MoghimiBMVC2016}, by reducing the feature dimension~\cite{gao16compact, KongF16}, or by aggregating higher order statistics~\cite{cui2017cvpr}.

\subsection{Datasets}\label{sec:datasets}
We present experiments on the Caltech-UCSD birds~\cite{WahCUB_200_2011}, Stanford cars~\cite{krause20133d}, and FGVC aircrafts~\cite{maji2013fine} fine-grained recognition datasets. The birds dataset provides 11,788 images across 200 species of birds with object bounding boxes and detailed part annotations. Birds are small, exhibit pose variations, and appear in clutter making recognition challenging. 
The Stanford cars dataset consists of 196 models categories with 16,185 images in total and provides object bounding boxes. The task is to recognize makes and models of cars, \eg, 2012 Tesla Model S or 2012 BMW M3 coupe. The aircrafts dataset contains 10,000 images of 100 different aircraft variants such as Boeing 737-300 and Boeing 737-400. Although these datasets provide different levels of annotations we only use category labels in our experiments.

\subsection{Models}\label{sec:models}
We experiment with B-CNNs based on the VGG-M~\cite{chatfield14return} and 16-layer VGG-D~\cite{simonyan14very} network. Following our earlier work~\cite{lin2015bilinear,lin2017bilinear} we resize the image to 448x448 pixels and extract the features from \emph{relu5} and \emph{relu5\_3} layers of the VGG-M and VGG-D network respectively.
\emph{For the aircraft dataset we found that a central crop of size 448x448 obtained from the image resized to 512x512 improves the accuracy of the baseline model~\cite{lin2015bilinear} by a few percentage points (\eg, 84.1\% $\rightarrow$ 86.9\% using the B-CNN model with VGG-D network)}. 
We use this cropping scheme for all experiments on the aircraft dataset. 
The bilinear pooling results in a 512x512 dimensional representation for both the networks. This is then passed through elementwise signed square-root normalization $(x \leftarrow sign(x)\sqrt{|x|})$ and $\ell_2$ normalization, followed by a linear SVM or logistic regression for classification.

For improved B-CNNs we apply the matrix logarithm or the matrix square-root normalization after bilinear pooling and add a small positive number to the diagonal to make the matrix strictly positive definite, as shown in Figure~\ref{fig:arch}. The maximum value of $\sigma_{max} \approx 10^6$ of the outer-product features and we find that $\epsilon=1$ is sufficient for to ensure that the matrix is positive definite to numerical precision. Optionally, the elementwise signed square-root normalization is applied after the matrix normalization steps. In all cases we apply $\ell_2$ normalization of the feature since this is just a global scaling and makes the selection of hyperparmeters in the classification layer consistent across models and datasets. The effect of various normalization schemes and their combinations are shown in Table~\ref{tab:norm}.

\textbf{Training and evaluation.} We follow same training and evaluation protocol~\cite{lin2015bilinear,lin2017bilinear} where for fine-tuning a k-way linear and softmax layer is added on top of the pre-trained networks. This layer is initialized using logistic regression after which the entire network is trained end-to-end using stochastic gradient descent with momentum for 50-100 epochs with a learning rate $\eta=0.001$. We perform left-right image flipping as data augmentation. 
After fine-tuning the last layer is replaced by k one-vs-rest linear SVMs with the hyperparameter $C=1$. 
At test time model is evaluated on an image and its flipped copy and their predictions are averaged. 
Both test-time flipping and SVM training can be ignored for the VGG-D networks for a negligible loss in accuracy. 
We report per-image accuracy on the test set provided in the dataset.

\subsection{Effect of feature normalization}\label{sec:norm}
\begin{wrapfigure}{r}{0.5\textwidth}
\vspace{-0.35in}
\centering
\includegraphics[valign=T,width=0.9\linewidth]{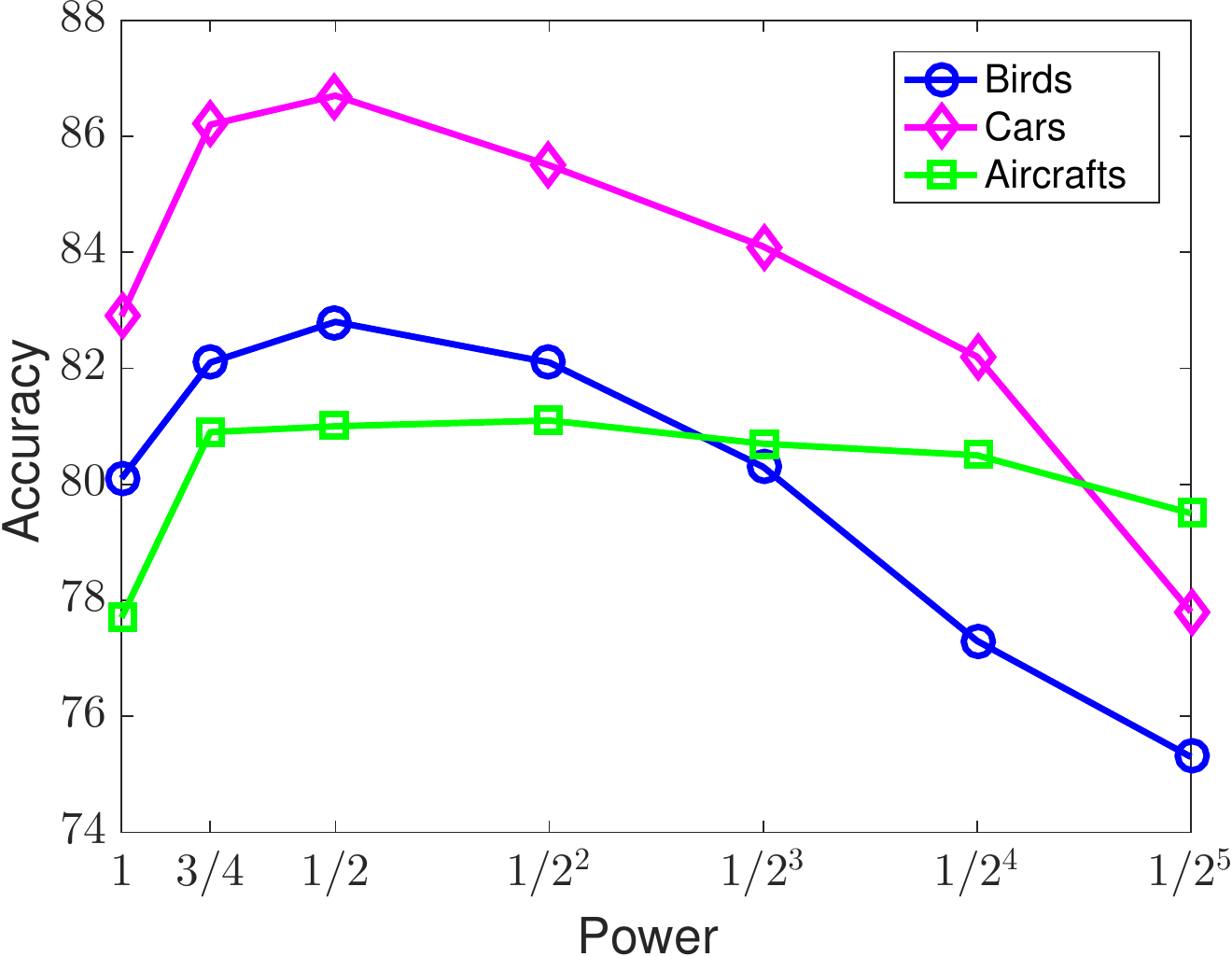}
\vspace{0.1in}
\caption{\label{fig:power}Accuracy \emph{vs.} the exponent $p$.}
\vspace{-0.1in}
\end{wrapfigure}

\paragraph{Effect of the exponent $p$ in the matrix power normalization $A^p$.} Figure~\ref{fig:power} shows that accuracy of non fine-tuned B-CNNs with the VGG-D network using power normalization $A^{p}$ for different values of the power $p$. The value $p=1$ corresponds to the baseline B-CNN accuracy~\cite{lin2015bilinear} where only elementwise signed square-root and $\ell_2$ normalization are applied. Note the improved results for aircrafts compared to~\cite{lin2015bilinear} due to the central cropping scheme described earlier. The plots show that the matrix square-root ($p=1/2$) works the best and outperforms the baseline B-CNN accuracy by a considerable margin.

\paragraph{Effect of combining various normalization schemes.} Table~\ref{tab:norm} shows the accuracy of non fine-tuned B-CNNs using various normalization schemes. The baseline model is shown as the row with only the $sgnsqrt(A)$ column checked. Matrix square-root or the matrix logarithm normalization alone does not always improve over elementwise signed square-root normalization. However, when combined, the improvements are significant suggesting that the two normalization schemes are complementary. Overall the combination of matrix square-root normalization is better than the matrix logarithm normalization. This is fortunate since the matrix square-root and its gradient can be accurately computed as described in Section~\ref{sec:matrix-functions}.

\begin{table}
\small
\centering
\begin{tabular}{c|c|c|c|c|c|c}
 & \multicolumn{3}{|c|}{Normalization} & \multicolumn{3}{|c}{Accuracy on dataset} \\
Network & $\log(A)$ & \sqA & $sgnsqrt(A)$ & Birds & Cars & Aircrafts \\
\hline
\multirow{5}{*}{VGG-M} &  	&  & \checkmark & 72.0 & 77.8 & 74.7 \\
								 & \checkmark &  & & 70.8 & 77.4 & 77.1 \\
								 &  & \checkmark &  & 70.3 & 76.8 & 75.0\\
& \checkmark & & \checkmark & 72.7 & 81.2 & \textbf{81.0}\\
&  & \checkmark & \checkmark & \textbf{76.3} & \textbf{83.4} & 80.7 \\
\hline
\multirow{5}{*}{VGG-D} &  	&  & \checkmark &  80.1 & 82.9  & 77.7 \\
								 & \checkmark &  &  & 77.9 & 79.8 & 78.7\\
								 &  & \checkmark &  & 80.6 & 82.3 & 78.7 \\
								 & \checkmark & & \checkmark & 81.1 & 85.1 & \textbf{81.4} \\
&  & \checkmark & \checkmark & \textbf{82.8} & \textbf{86.7} & 80.9 \\
\end{tabular}
\vspace{0.1in}
\caption{\label{tab:norm} Accuracy of B-CNNs with different normalization schemes with non fine-tuned networks. The best results are obtained with matrix-square root followed by element-wise signed square-root normalization. Notably the matrix square-root is better than the matrix logarithm normalization on most datasets for both networks.}
\end{table}

\subsection{Effect of network fine-tuning}\label{sec:fine-tuning}
We perform fine-tuning of the network using the matrix square-root layer in combination with elementwise square-root layer. Table~\ref{tab:fine-tuning} shows the results with fine-tuning B-CNNs with the VGG-M and VGG-D networks.
For these experiments the gradients were computed using the Lyapunov technique described in Section~\ref{sec:matrix-functions}. 
Training with SVD-based gradients led to a slightly worse performance, details of which are described in the next section.
The matrix square-root normalization remains useful after fine-tuning and results in a \textbf{2-3\%} improvement on average across the fine-grained datasets for both networks. The improvements are especially large for the VGG-M network.

\begin{table}
\centering
\begin{tabular}{c|c|c|cc|cc|cc} & \multicolumn{2}{|c|}{Normalization} & \multicolumn{6}{|c}{Accuracy on dataset} \\
Network & \sqA & $sgnsqrt(A)$ & \multicolumn{2}{|c|}{Birds} & \multicolumn{2}{|c|}{Cars} & \multicolumn{2}{|c}{Aircrafts} \\
\hline
\multirow{2}{*}{VGG-M} &  & \checkmark & 72.0 & 78.1 & 77.8 & 86.5 & 74.7 & 81.3 \\
& \checkmark & \checkmark & 76.3 & \textbf{81.3} & 83.4 & \textbf{88.5} & 80.7 & \textbf{84.0} \\
\hline
\multirow{2}{*}{VGG-D} &  & \checkmark & 80.1 & 84.0 & 82.9 & 90.6 & 77.7 & 86.9 \\
& \checkmark & \checkmark & 82.8 & \textbf{85.8} & 86.7 & \textbf{92.0} & 80.9 & \textbf{88.5}\\
\end{tabular}
\vspace{0.1in}
\caption{\label{tab:fine-tuning} For each dataset the accuracy before and after end-to-end fine-tuning of the networks are shown on the left and right column respectively. Matrix square-root normalization provides consistent improvements in accuracy over the baseline across all datasets.}
\end{table}

\subsection{Are exact computations necessary?}\label{sec:precision} 
Despite the improvement it offers a drawback of the matrix square-root is that computing the SVD is relatively slow and lacks batch-mode implementations. For the VGG-D network the computing the SVD of a 512x512 matrix takes about 22 milliseconds on a NVIDIA Titan X GPU, which is comparable to the rest of the network evaluation. 
Instead of computing the matrix square-root accurately using SVD, one can compute it approximately using a few iterations of the modified Denman-Beavers iterations described in Section~\ref{sec:matrix-functions}.
Table~\ref{tab:newton-iterations} shows that accuracy on the final classification task and time taken as a function of number of iterations. 
As few as 5 iterations are sufficient for matching the accuracy of the SVD method while being 5$\times$ faster. 
Surprisingly, even a \emph{single} iteration provides non-trivial improvements over the baseline model (0 iterations) and takes less than 1 millisecond to evaluate.
Although we didn't implement it, the method can be made faster using a batch-mode version of these iterations. 
With these iterative methods matrix-normalization layers are no longer the bottleneck in network evaluation.

\begin{table}[h]
\centering
\begin{tabular}{c|c|c|c|c|c|c||c|c|c}
\multicolumn{7}{c||}{Forward} & \multicolumn{3}{c}{Backward} \\
Iterations & 0 & 1 & 5 & 10 & 20 & SVD & LYAP & SVD & Faster\\
\hline
Birds & 80.1 & 81.7 & 83.0 & 82.9 & 82.8 & 82.8 & 85.8 & 85.5 &  85.3\\
Cars & 82.9 & 85.0 & 87.0 & 86.8  & 86.7 & 86.7 & 92.0 & 91.8 &  91.4 \\
Aircrafts & 77.7 & 79.5 & 81.3 & 81.1 & 80.9 & 80.9 & 88.5 & 87.8 & 86.8 \\
\hline
Time & 0 & 1ms & 4ms & 6ms & 11ms & 22ms & - & - & - 
\end{tabular} 
\vspace{0.2in}
\caption{\label{tab:newton-iterations} On the left is the effect of number of Newton iterations for computing the matrix square-root on the speed and accuracy of the network. On the right is the accuracy obtained using various gradient computation techniques in the backward step. The VGG-D network is used for this comparison.}
\end{table}

Table~\ref{tab:newton-iterations} also shows the accuracy of fine-tuning with various gradient schemes for the matrix square-root.
The time taken for backward computations (both LYAP and SVD) are negligible given the SVD decomposition computed in the forward step and hence are not shown in the table.
A faster scheme where the matrix square-root layer is ignored during fine-tuning is worse, but in most cases outperforms the fine-tuned baseline B-CNN model (Table~\ref{tab:fine-tuning}). Although we found that SVD gradients are orders of magnitude less precise than Lyapunov gradients, the loss in accuracy after fine-tuning is negligible. 

Our attempts at fine-tuning the network with matrix-logarithm using SVD-based gradients were not very successful, even with double precision arithmetic. On the other hand, all the experiments with matrix square-root were done with single precision arithmetic.
This suggests that the numerical issues are partly due to the logarithm scaling of the eigenvalues.

%

\subsection{Comparison with prior work}\label{sec:prior-work}
Since its introduction the B-CNN architecture has been improved in a number of ways. Table~\ref{table:comp} compares the Improved B-CNN with those extensions along with other best performing methods. The Compact Bilinear Pooling (CBP)~\cite{gao16compact} approach applies Tensor Sketch~\cite{pham2013fast} to approximate the second-order statistics and reduces the feature dimension by two orders of magnitude achieving 84\% accuracy on birds dataset. LRBP~\cite{KongF16} applies low-rank approximation to the covariance matrix further reducing the computational complexity obtaining 84.2\% accuracy on birds. BoostCNN~\cite{MoghimiBMVC2016} boosts multiple B-CNNs trained at multiple scales and achieves the best results on birds (86.2\%) and aircrafts (88.5\%) without requiring additional annotations at training time. However, their approach is one to two orders of magnitude slower. The recently-proposed Kernel Pooling (KP)~\cite{cui2017cvpr} approach aggregates higher-order statistics by iteratively applying the Tensor Sketch compression to the features and achieves 86.2\% on birds. Improved B-CNN achieves comparable results to BoostCNN and KP for a small increase in computational cost compared to the baseline architecture. These approaches are complementary and may be combined for added benefits.

\begin{table}[t]
\centering
\small
\begin{tabular}{r|r|c|c|c|c|c}
 & & \multicolumn{2}{|c|}{Annotation} & \multicolumn{3}{|c}{Accuracy on dataset} \\
 \hline
Method & Network & Train & Test & Birds & Cars & Aircrafts \\
\hline
B-CNN  & VGG-M & - & -  & 78.1 & 86.5 & 81.3 \\
B-CNN  & VGG-D & - & -  & 84.0 & 90.6 & 86.9 \\
B-CNN  & VGG-M + VGG-D & - & - & 84.1 & 91.3 & 86.6 \\
\hline
\textbf{Improved B-CNN} & VGG-M & - & - & 81.3 & 88.5 & 84.0 \\
\textbf{Improved B-CNN} & VGG-D & - & - & \textbf{85.8} & \textbf{92.0} & \textbf{88.5} \\
\hline
STNs~\cite{jaderberg15spatial} & Inception-BN & - & - & 84.1 & - & - \\
BoostCNN~\cite{MoghimiBMVC2016} & B-CNN & - & - & \textbf{86.2} & 92.1 & \textbf{88.5} \\
Krause \etal~\cite{krause2015fine}& VGG-D & Box & - & 82.0 & \textbf{92.6} & - \\
PD+SWFV-CNN~\cite{Zhang2016CVPRpdfs} & VGG-D & - & - & 83.6 & - & - \\
PD+FC+SWFV-CNN~\cite{Zhang2016CVPRpdfs} & VGG-D & - & - & 84.5 & - & - \\
CBP~\cite{gao16compact} & VGG-D & - & - & 84.0 & - & - \\
LRBP~\cite{KongF16} & VGG-D & - & - & 84.2 & 90.9 & 87.3 \\
KP~\cite{cui2017cvpr} & VGG-D & - & - & \textbf{86.2} & 92.4 & 86.9 \\
\end{tabular}
\vspace{0.1in}
\caption{\label{table:comp}Comparison with the prior work. The second column shows the basic networks used for each method, and the third and fourth columns show the extra annotations used during train and test. Our improved B-CNN consistently outperforms the baseline B-CNN architecture~\cite{lin2015bilinear} by a significant margin and is comparable to the state-of-the-art.}
\end{table}

\section{Conclusion}
\urlstyle{rm}
We propose an improved B-CNN architecture that uses matrix-normalization layers, offers complementary benefits to elementwise normalization layers, and leads consistent improvements in accuracy over the baseline model.
The matrix square-root normalization outperforms the matrix logarithm normalization when combined with elementwise square-root normalization for most of our experiments.
Moreover, the matrix square-root can be computed efficiently on a GPU using few Newton iterations and allows accurate gradient computations via a Lyapunov equation.
Interestingly, even a single iteration provides improvements over the baseline architecture while adding negligible cost to the forward computation.
For future work we will explore if computing the gradients by unrolling the iteration is sufficiently accurate for training. 
This can be faster and would allow the subsequent layers to adapt to the errors of the iterative method. The source code and pre-trained models will be made available on the project page: \url{http://vis-www.cs.umass.edu/bcnn}.
\paragraph{Acknowledgement:}
This research was supported in part by the NSF grants IIS-1617917 and ABI-1661259, and a faculty gift from Facebook. The experiments were performed using high performance 
computing equipment obtained under a grant from the Collaborative R\&D Fund 
managed by the Massachusetts Tech Collaborative and GPUs donated by NVIDIA.

\bibliography{bibliography}
\end{document}